\documentclass{article}

\PassOptionsToPackage{numbers, compress}{natbib}
\usepackage[preprint]{tackling_climate_workshop_style}

\usepackage[utf8]{inputenc} 
\usepackage[T1]{fontenc}    
\usepackage{hyperref}       
\usepackage{url}            
\usepackage{booktabs}       
\usepackage{amsfonts}       
\usepackage{nicefrac}       
\usepackage{microtype}      
\usepackage{caption}
\usepackage{color}
\usepackage{subcaption}
\usepackage{graphicx}
\usepackage[toc,page]{appendix}
\usepackage{paralist}
\usepackage{float}
\usepackage{amsmath}

\title{Generative Modeling of High-resolution Global Precipitation Forecasts}

\author{%
  James Duncan\\
  UC Berkeley\\
  Berkeley, CA 94720 \\
  \texttt{jpduncan@berkeley.edu} \\
  \And
  Shashank Subramanian\\
  Lawrence Berkeley National Laboratory\\
  Berkeley, CA 94720 \\
  \texttt{shashanksubramanian@lbl.gov} \\
  \And
  Peter Harrington\\
  Lawrence Berkeley National Laboratory\\
  Berkeley, CA 94720 \\
  \texttt{pharrington@lbl.gov} \\
}

\begin{document}

\maketitle

\begin{abstract}
    Forecasting global precipitation patterns and, in particular, extreme precipitation events is of critical importance to preparing for and adapting to climate change. Making accurate high-resolution precipitation forecasts using traditional physical models remains a major challenge in operational weather forecasting as they incur substantial computational costs and struggle to achieve sufficient forecast skill. Recently, deep-learning-based models have shown great promise in closing the gap with numerical weather prediction (NWP) models in terms of precipitation forecast skill, opening up exciting new avenues for precipitation modeling. However, it is challenging for these deep learning models to fully resolve the fine-scale structures of precipitation phenomena and adequately characterize the extremes of the long-tailed precipitation distribution. In this work, we present several improvements to the architecture and training process of a current state-of-the art deep learning precipitation model (FourCastNet) using a novel generative adversarial network (GAN) to better capture fine scales and extremes. Our improvements achieve superior performance in capturing the extreme percentiles of global precipitation, while comparable to state-of-the-art NWP models in terms of forecast skill at 1--2 day lead times. Together, these improvements set a new state-of-the-art in global precipitation forecasting.
    

\end{abstract}

\section{Introduction}

Precipitation is a fundamental climate phenomenon with major impact on crucial infrastructure such as food, water, energy, transportation, and health systems \citep{IPCC22}.
On timescales relevant to weather prediction, making accurate forecasts of precipitation and extreme events is critical to the planning and management of these systems, along with disaster preparedness for extreme precipitation events, which are greatly amplified by climate change \citep{Rothfusz18}. Unfortunately, extreme precipitation remains one of the most challenging atmospheric phenomena to forecast accurately, due to high spatiotemporal variability and the myriad of complex multi-scale, multi-phase processes that govern its behavior \citep{Chen21, Tapiador19, Yano18}.

State-of-the-art NWP models, such as the Integrated Forecast System\footnote{https://www.ecmwf.int/en/forecasts/documentation-and-support} (IFS), produce operational forecasts by combining physics-based PDE solvers with assimilated observations from a variety of sources. The complexity of moisture physics requires many parameterizations in these models for processes like turbulent mixing, convection, subgrid clouds and microphysics \citep{IFS20}, and such parameterizations can lead to large biases in NWP precipitation forecasts \citep{Lavers21}. As a result, global precipitation forecasts generally achieve inadequate forecast skill \citep{Yano18} and, hence, there has been an increasing interest in fully data-driven solutions, primarily using deep learning, in recent years. 

\begin{figure}
  \centering
  \includegraphics[width=\linewidth]{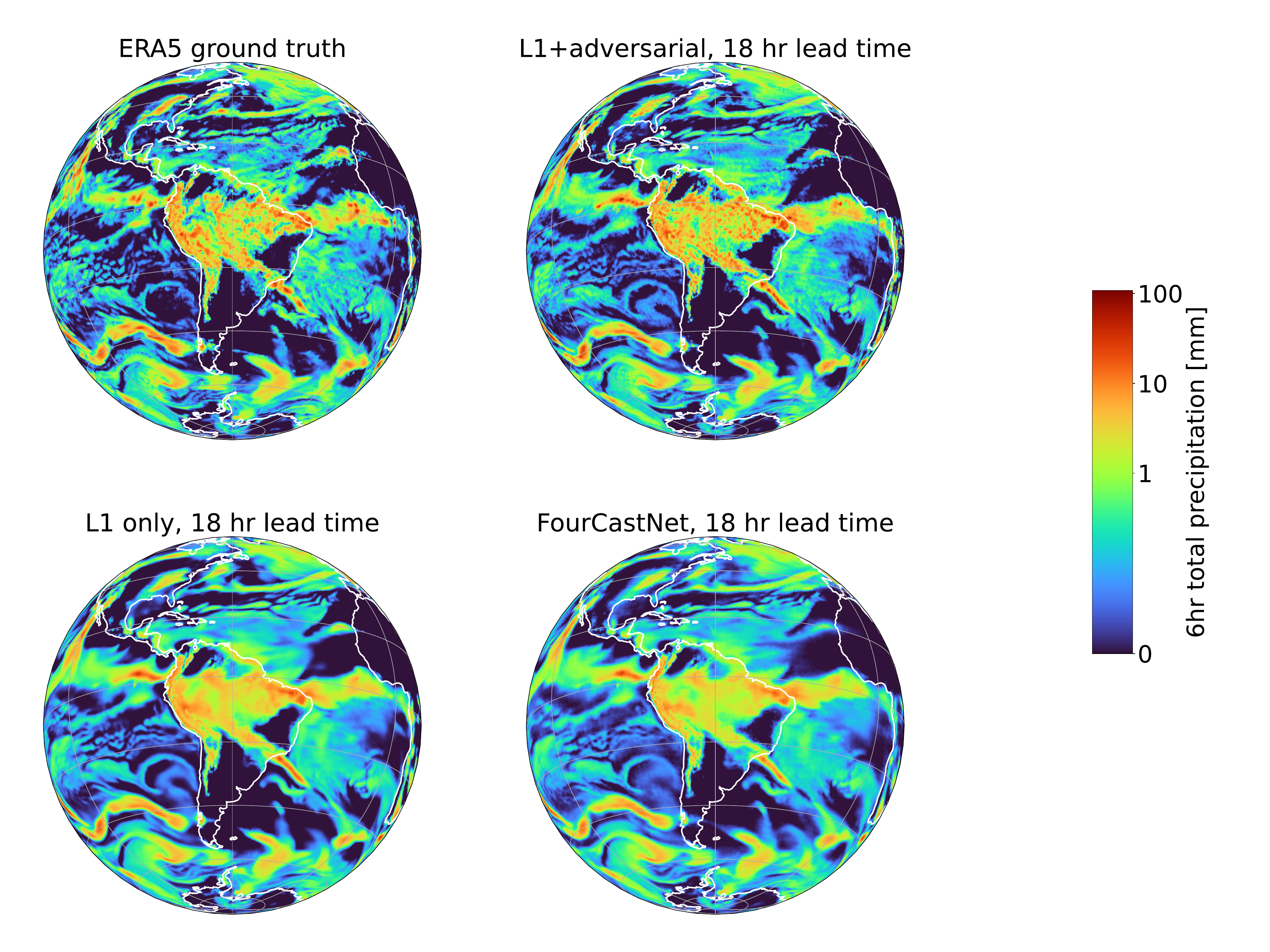}
  \caption{Visualization of precipitation forecasts at 18hr forecast lead time, comparing fine-scale features between FourCastNet, our $\mathcal{L}_1$-only model, our adversarial+$\mathcal{L}_1$ model, and the ground truth ERA5 over South America. We observe that the adversarially trained model shows finer-scale structures and matches the ground truth more accurately. }
  \label{fig:viz}
\end{figure}

Data-driven models can be orders of magnitude faster with the potential to learn complex parameterizations between input and output function spaces directly from data, reducing model bias.
With such models, major advances have been made in the area of precipitation ``nowcasting'', where forecasts are made over limited spatial regions with lead times on the order of minutes to hours. Deep learning models trained directly on radar and/or satellite observations now outperform traditional methods for nowcasting \citep{DMNowcast21, MSNowcast21, MetNetNowCast22}. However, until recently, there has been limited progress for models predicting precipitation at larger spatiotemporal scales (e.g., over the full globe up to days in advance), mainly due to computational limitations on resolution \citep{Rasp21}. FourCastNet \citep{pathak2022fourcastnet} is the first deep-learning-based global weather model running at $\sim$30km scale, which outperforms IFS in terms of precipitation forecast skill up to $\sim$2 day lead times and is the current state-of-the-art.
However, despite using a dedicated network just for precipitation due to its unique challenges, FourCastNet predictions still lack fine-scale details (see Figure \ref{fig:viz}) and thus underestimate the extreme percentiles of the precipitation distribution.

In this work, we aim to overcome some of these limitations using generative models to advance the state-of-the-art in deep learning-based precipitation forecasts. In particular,
our contributions are as follows:
\begin{inparaenum}[(i)]
    \item we apply a state-of-the-art generative adversarial network \citep{Jiang20} that integrates multi-scale semantic structure and style information, allowing us to synthesize physically realistic fine-scale precipitation features; 
    \item we show that capturing fine-scale phenomena leads to improved predictions of extreme precipitation while also preserving forecast skill, attaining comparable skill at 1--2 lead day times with respect to IFS.
\end{inparaenum}
\section{Methods}
\subsection{Dataset}
We replicate the data preparation pipeline of the original FourCastNet precipitation model \citep{pathak2022fourcastnet}, relying on the European Center for Medium-Range Weather Forecasting (ECMWF) global reanalysis dataset ERA5 \citep{ERA5}, which combines archived atmospheric observations with physical model outputs. Following FourCastNet's time steps of length 6 hours, we model the 6hr accumulated total precipitation $TP$ (this also makes for easier comparisons against IFS, which archives $TP$ forecasts in 6 hourly accumulations as well). A sample snapshot of $TP$, in log-normalized units for easy visualization of fine-scale details, is shown in the bottom-right inset of Figure \ref{fig:viz}. We use the years 1979-2015 and 2016-2017 as training and validation sets, respectively, and set aside 2018 as a test set. We refer the reader to the original FourCastNet paper \citep{pathak2022fourcastnet} for further details on the dataset.

\subsection{Model}
\begin{figure}
  \centering
  \includegraphics[width=.82\linewidth]{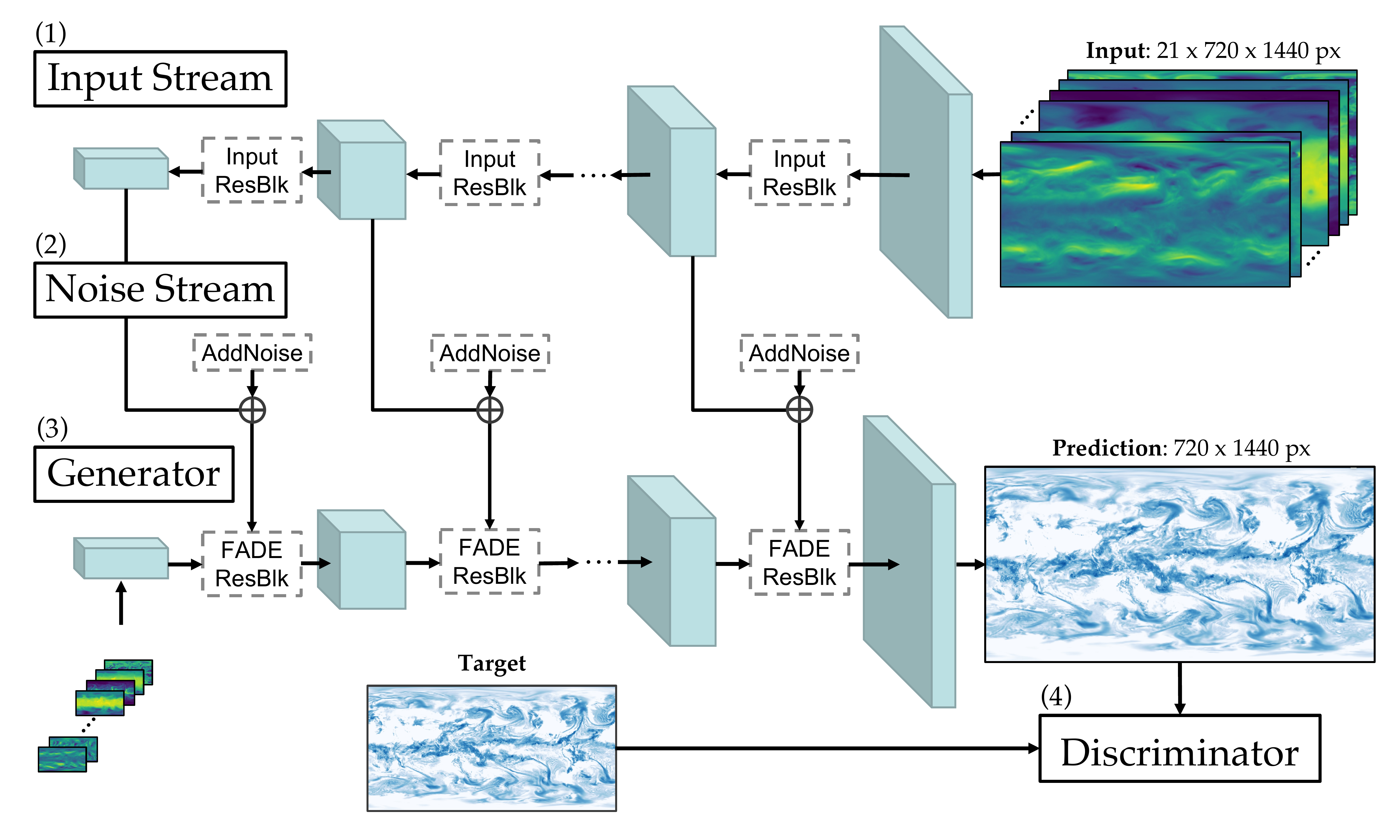}
  \caption{Training architecture of the \emph{adversarial$+\mathcal{L}_1$} model, a modified version of TSIT \citep{Jiang20}. The model learns multi-scale representations of the input, perturbed by injected noise, and generates stochastic images with fine-scale detail promoted by a multi-scale discriminator.
  }
  \label{fig:arch}
\end{figure}

Given the success of adversarial learning for high-resolution precipitation models in localized regions \citep{DMNowcast21, leinonen2020stochastic, price2022increasing}, we explore the utility of conditional generative adversarial networks (cGANs) for modeling $TP$ over the entire globe using prognostic atmospheric variables as conditional input for operational diagnosis. In particular, we adopt the TSIT architecture \citep{Jiang20} for our task, due to its success in varied image tasks and ability to flexibly condition intermediate features at a variety of scales in the generator. TSIT's generator $G$ employs symmetric downsampling and upsampling paths, fusing features from the downsampling path via feature-wise adaptive denormalization, and uses multi-scale patch-based discriminators \citep{wang2018high} for adversarial training. We take a constant latitude embedding channel and the 20 atmospheric variables output by FourCastNet (corresponding to various different prognostic variables such as wind velocities, total column water vapor and temperature at different pressure levels) 
as input to the network, as shown in Figure \ref{fig:arch}. Randomness is injected 
at intermediate scales via elementwise additive noise, and 
we can thus generate ``zero-shot'' ensembles for probabilistic forecasting given a single input, which we explore in Appendix \ref{app:arch}. We refer the reader to \citep{Jiang20} for additional details on the TSIT architecture, and list hyperparameters in Appendix \ref{app:arch}, together with the particulars of our setup. 
In this work, we focus on three model versions for demonstrating the impact of the adversarial training: \begin{inparaenum}[(i)] \item \emph{FourCastNet}: the precipitation baseline model from \citep{pathak2022fourcastnet}, \item \emph{$\mathcal{L}_1$-only}: TSIT model with a simple (non-adversarial) $\mathcal{L}_1$ loss, and \item \emph{adversarial$+\mathcal{L}_1$}: TSIT model with adversarial training. \end{inparaenum}

\section{Results}
\begin{figure}
    \centering
    \begin{subfigure}[t]{0.49\textwidth}
        \centering
        \includegraphics[width=\linewidth]{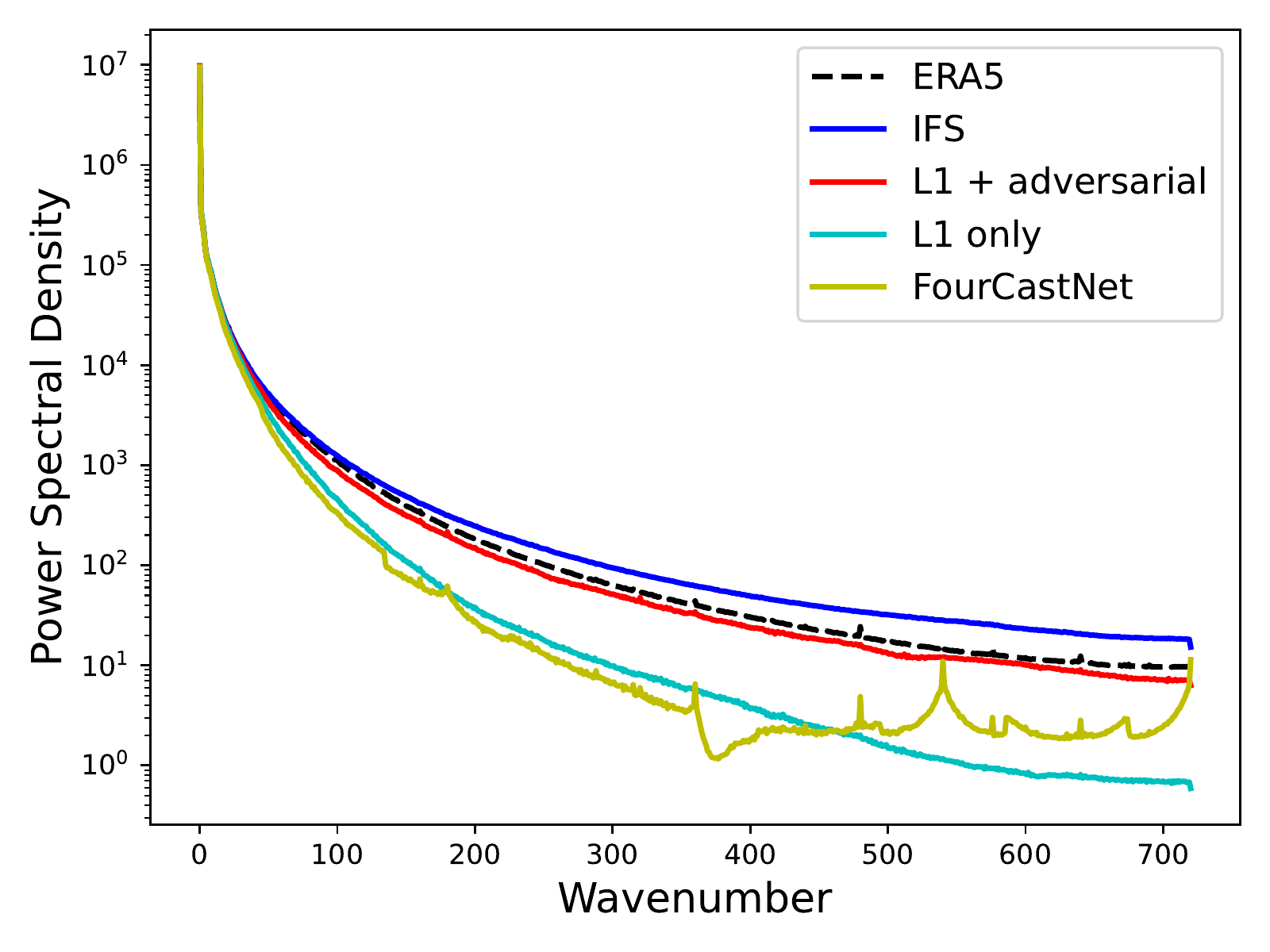} 
    \end{subfigure}
    \hfill
    \begin{subfigure}[t]{0.49\textwidth}
        \centering
        \includegraphics[width=\linewidth]{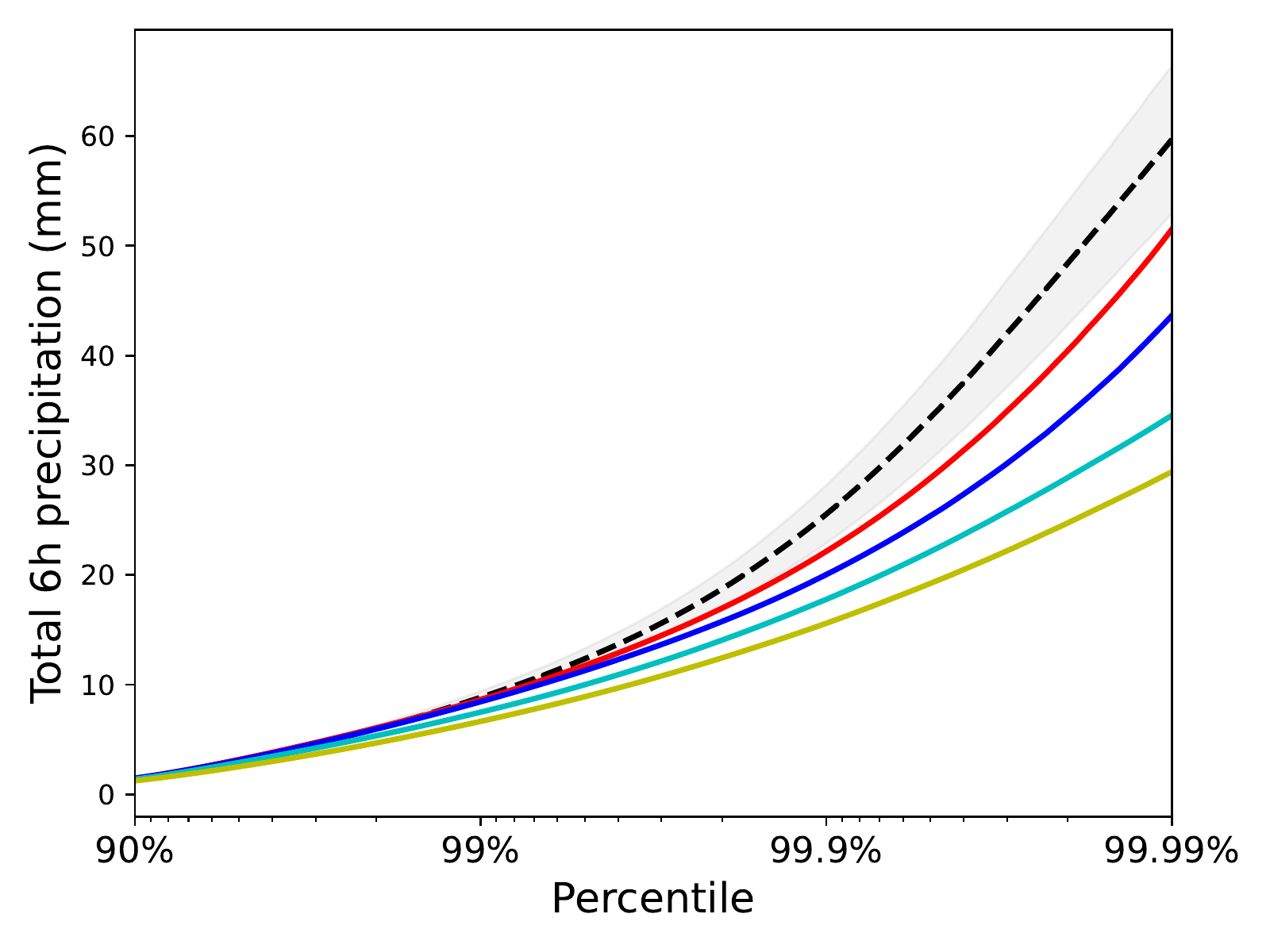} 
    \end{subfigure}
    \caption{Analysis of precipitation power spectrum (left) and extreme percentiles (right) at 18hr lead time. Results are averaged over 178 initial conditions spread uniformly across 2018, and the gray shaded region shows the ERA5 variability in percentiles over initial conditions. 
    }
    \label{fig:stats}
\end{figure}

We observe that, qualitatively, the adversarial training procedure clearly improves the perceptual realism and fine-scale detail represented in precipitation forecasts, as presented in the visualizations in Figure \ref{fig:viz} which compares $TP$ from the 2018 test set over South America at 18hr lead time. We confirm these results quantitatively by computing a 1D power spectrum of the $TP$ forecasts along the East-West direction, again at 18hr lead time, averaged over several initial conditions in the test set\footnote{The initial conditions are spaced apart by 2 days as a rough estimate of the temporal decorrelation}. The results are plotted in Figure \ref{fig:stats}a, where we find that that the adversarial learning framework effectively matches the the ground truth spectrum, specifically at higher wavenumbers (and hence finer spatial scales), outperforming all other models.
Beyond assessing fine scales, the general task of identifying metrics for evaluating precipitation forecast quality is itself a challenge \citep{Tapiador19, Leung22}. For example, Anomaly Correlation Coefficient (ACC) is widely used to assess forecast skill, but it is biased towards larger scales, so smoothing functions can artificially inflate this metric (we illustrate this with simple gaussian blur in Appendix \ref{app:additional_stats}). Furthermore, ACC is generally insensitive to extremes which is a key quantity to capture in fields such as precipitation.
While our model is comparable to the IFS in terms of ACC on 1-2 day lead times (see Appendix \ref{app:additional_stats}), we emphasize here the model's ability to more faithfully capture the extreme $TP$ values that are of interest to weather and climate stakeholders.
We plot the extreme percentiles of our forecasts, again at 18hr lead time, in Figure \ref{fig:stats}b, binning logarithmically to emphasize the tail of the distribution \citep{Fildier21} and averaging over the 2018 test set. Clearly, the model better replicates the long tail of the $TP$ distribution, nearly matching the ground truth percentiles within the variability over initial conditions in the test set, and outperforms IFS in this regard as well. We show additional results on the $TP$ distribution in Appendix \ref{app:additional_stats}.

\section{Conclusions}
Deep-learning-based models like FourCastNet show great promise in data-driven forecasting of precipitation ($TP$), but exhibit smooth features and underestimate extreme events. In this work, we demonstrate the ability of GAN-based models to resolve small-scale details and synthesize physically realistic extreme values in the tail of the $TP$ distribution, outperforming IFS (a leading NWP model) on both counts while retaining competitive forecast skill in terms of ACC. Our model sets a new state-of-the-art in global $TP$ forecasting, and can be complementary to existing localized, short-timescale, high-resolution precipitation models which require initial or boundary conditions from NWP (or similar) models as input.

\begin{ack}
The authors would like to thank Jaideep Pathak and Karthik Kashinath for helpful discussions. This research used resources of the National Energy Research Scientific Computing Center (NERSC), a U.S. Department of Energy Office of Science User Facility located at Lawrence Berkeley National Laboratory, operated under Contract No. DE-AC02-05CH11231
\end{ack}

\small
\bibliographystyle{abbrvnat}
\bibliography{refs}

\normalsize
\begin{appendices}

\section{Additional training details and probabilistic inference}
\label{app:arch}

\subsection{Hyperparameters \& architecture details}

The \emph{$\mathcal{L}_1$-only} and \emph{adversarial$+\mathcal{L}_1$} models both use a modified version of the architecture described in \citep{Jiang20}. 
We list hyperparameters in Table~\ref{tab:hp}, and refer the reader to \citep{Jiang20} for further details on the hyperparameters.

\begin{table}[h]
    \caption{Hyperparameters for \emph{adversarial$+\mathcal{L}_1$} model, with any differences in the case of the $\mathcal{L}_1$-only model listed between square brackets.}
    \label{tab:hp}
    \centering
    \begin{tabular}{ll}
    \toprule
    Hyperparameter & Value for adversarial training [$\mathcal{L}_1$-only, if different] \\
    \midrule
    Global batch size & 64 \\
    Optimizer & Adam \\
    Initial learning rate & $2.5 \times 10^{-4}$ [$6.5 \times 10^{-4}$] \\
    LR schedule num. epochs & 15 constant + 6 linear decay [14 + 0] \\
    $\beta_1, \beta_2$ & 0.85, 0.95 \\
    $\lambda_{\text{feat}}$ & 0.5 [N/A] \\
    Number of upsampling blocks & 8 \\
    Input stream normalization & Spectral instance \\
    Generator normalization & FADE + spectral batch \\
    Discriminator normalization & Spectral instance [N/A] \\
    Multi-scale discriminators & 4 \\
    Discriminator layers & 6 \\
    Padding & Zeros \\
    \bottomrule
    \end{tabular}
\end{table}

During training of the \emph{adversarial$+\mathcal{L}_1$} model, a fixed latitude-embedding channel is appended to the 20 prognostic input variables used in \citep{pathak2022fourcastnet} and passed to the input stream at full-resolution (720px $\times$ 1440px).
The entry point (layer $k=0$) to the input stream is a standard convolutional residual layer~\citep{He_2016_CVPR} with 64 output channels (pictured in the upper right of Figure~\ref{fig:arch}).
Next, these intermediate features are downsampled to 512px $\times$ 1024px using nearest neighbor interpolation and passed to the next residual layer of similar design, doubling the number of output channels to 128. 
Starting here, the output features are stored for multi-scale synthesis in the generator at a later stage, with increasing depth of field until the final layer ($k=8$).
The input stream uses downsampling at the beginning of each residual block, and the channel dimension is doubled after each residual block up to $d=1024$ at layer $(k=4)$,
resulting in a final representation with dimensions 4px $\times$ 8px $\times$ 1024 channels.

The generator mirrors the input stream, but travels in the opposite direction, from global to local representations. 
First, we downsample the 21 input fields to the same spatial dimensions as the final representation from the input stream.
Then, a fully-connected layer takes in the downsampled input and returns the starting set of 1024 features (lower left of Figure~\ref{fig:arch}), matching the full dimensions of the corresponding input stream features.
These features then pass through the first of eight feature-adaptive denomoralization (FADE) layers, integrating multi-scale semantic information from the input stream at the corresponding points in the generator.

To further promote diversity in the generator's outputs, the noise stream draws random gaussian noise which it adaptively scales in a feature-specific, multi-scale manner. 
This noise is added to the input stream's features prior to synthesis with the current output representation in each FADE layer.
The generator's intermediate features are then upsampled before passing on to the next layer.
Once the generator reaches the penultimate resolution of 512px $\times$ 1024px by 64 channels, it upsamples to the full resolution, applies Leaky ReLU with negative slope of 0.2, passes the features through a final convolutional layer for single-channel output, and applies the ReLU activation function to enforce precipitation non-negativity.

The multi-scale discriminator consists of four separate discriminators, each processing the same input features at a different spatial scale. One discriminator operates at full-resolution, while the others process inputs which are downsampled by progressive factors of 2 along the spatial dimensions.
Each discriminator takes as input the real or predicted $TP$ field, along with the input atmospheric variables fed to the generator (which get concatenated along the channel dimension) for improved training stability.
Each discriminator's output is a 2D grid of predictions classifying patches of the input as real or fake. This``patchGAN'' approach helps the model focus more on local texture and fine-scale details \citep{wang2018high}.

In addition to the adversarial loss term using the discriminators' outputs, a feature-matching loss \citep{salimans2016improved} term is used to improve training stability for the generator.
We refer the reader to \citep{Jiang20, wang2018high} for further specifics on the FADE and multi-scale patch-based discriminator architectures.
The architecture of the \emph{$\mathcal{L}_1$-only}  model resembles that of \emph{adversarial$+\mathcal{L}_1$} as pictured in Figure~\ref{fig:arch}, but with the noise stream and multi-scale discriminator components removed.

\subsection{Probabilistic ensemble forecasts}

\begin{figure}[H]
  \centering
  \includegraphics[width=\linewidth]{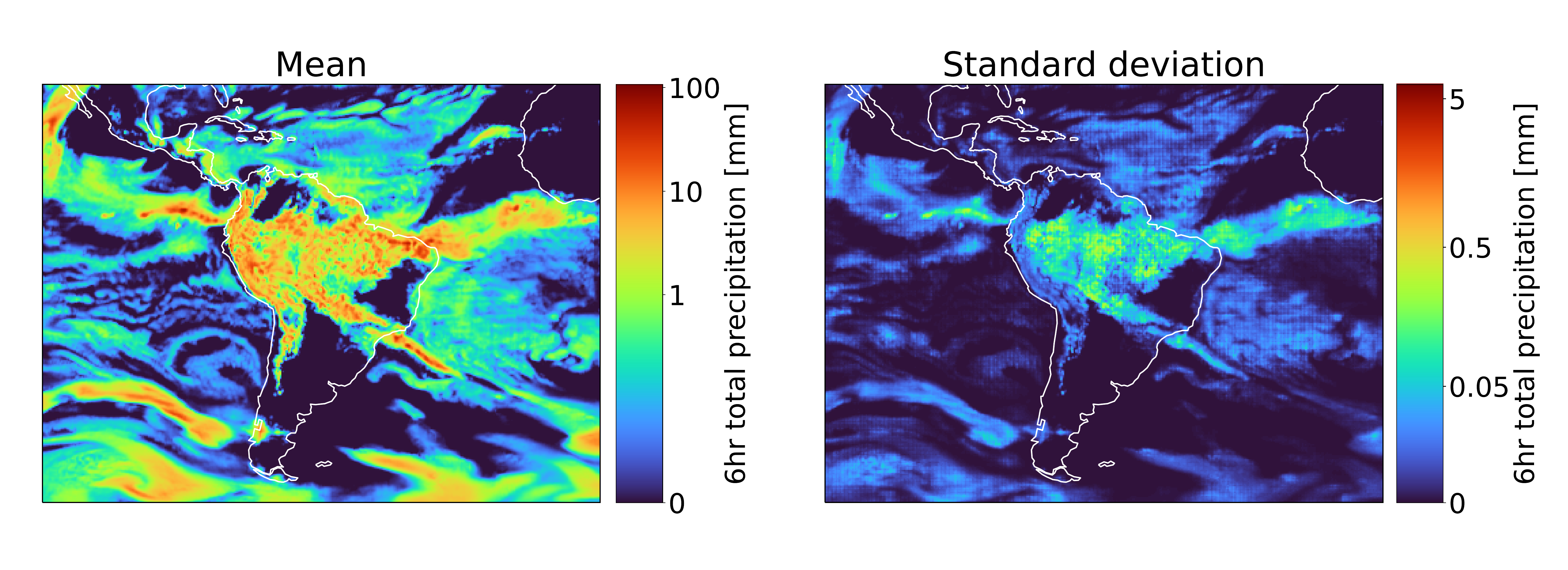}
  \caption{Through the introduction of randomness in the form of added noise between the input stream and generator, probabilistic forecasts are created by repeated inference runs with the same inputs, leading to an ensemble of unique outputs. We can then use the mean of this set of stochastic outputs as a final ensemble prediction for a given timepoint (left panel), and analyze the ensemble's standard error field to quantify forecast uncertainty at particular geographic locations (right panel).}
  \label{fig:ensemble_inf}
\end{figure}

Through the injection of adaptive randomness by the noise stream, outputs from the \emph{adversarial$+\mathcal{L}_1$} model are non-deterministic. Hence, we can output a number of repeated predictions and form a probabilistic ensemble forecast for a single timepoint, as depicted in Figure~\ref{fig:ensemble_inf}, even if we use a single control forecast for the other atmospheric variables input to the generator.

An alternate method for attaining multi-modal predictions, first introduced in the TSIT paper \citep{Jiang20}, is a variational auto-encoder (VAE) approach in which the bottom level of the generator is a latent space given by an auxiliary encoder and shaped by an additional KL-divergence loss. In  preliminary tests we found that this was unable to attain sufficient variability in predictions, and instead opted for directly perturbing the generator synthesis pipeline with our additive noise stream. However, further investigation is needed to determine the optimal method for multi-modal predictions in this application.

We present an example ensemble forecast in Figure~\ref{fig:ensemble_inf}, plotting mean and standard deviation per gridbox over $n=100$ ensemble members. Qualitatively, we observe highest variance over challenging locations like coastlines and mountainous regions, which aligns with our expectations as such areas exhibit high spatiotemporal variability and sharp, fine-scale features for $TP$. In future work we hope to quantitatively assess the quality of such ensemble forecasts, as rapid probabilistic forecasting from such a model would be a major step forward.

\section{Additional metrics and analysis}
\label{app:additional_stats}

\begin{figure}[H]
  \centering
  \includegraphics[width=0.6\linewidth]{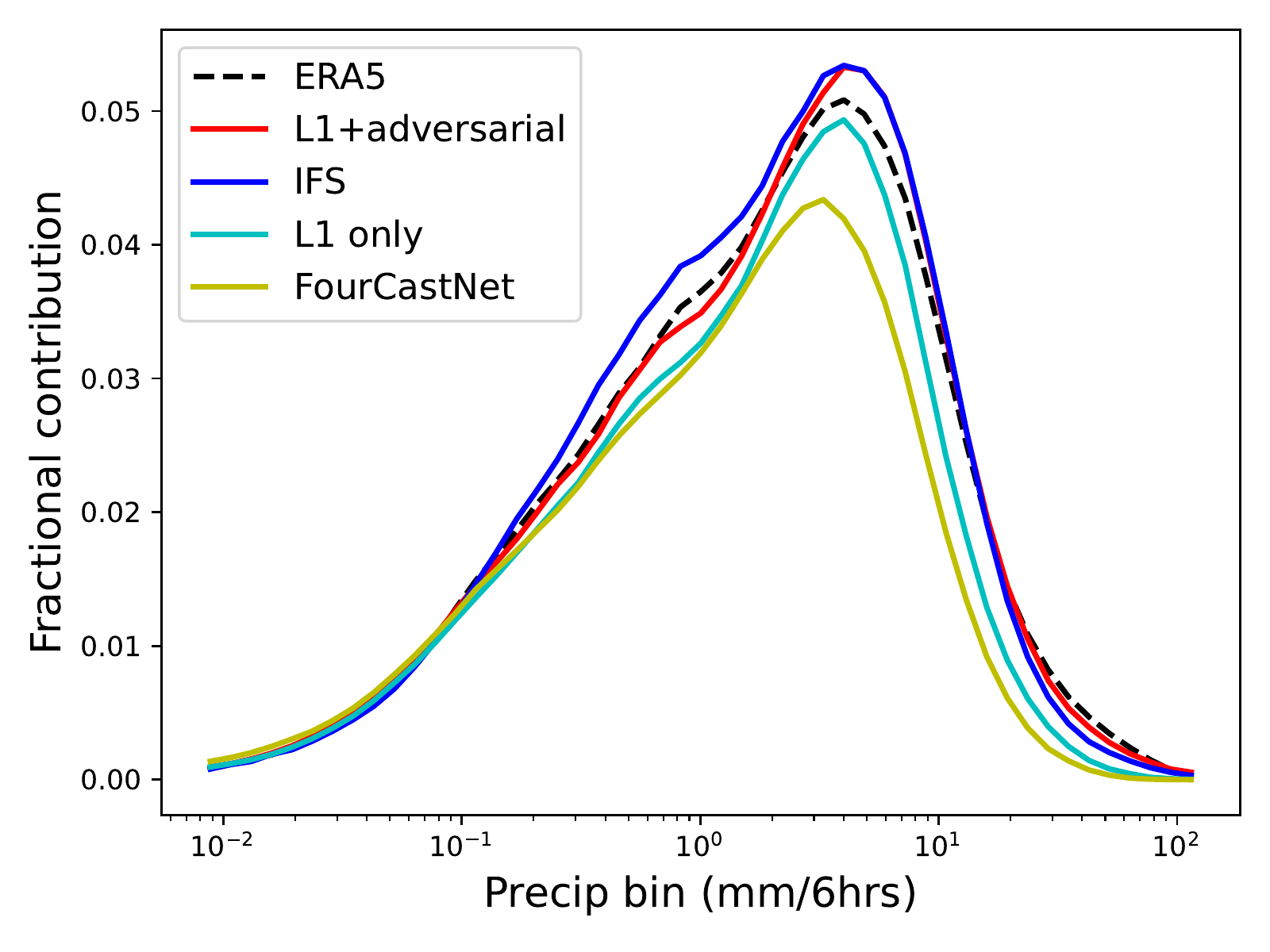}
  \caption{Fractional contributions of precipitation rates at 18hr lead time compared over the models in question and the ERA5 ground truth. Our model more faithfully captures the ground truth in general, but overestimates the central mode, similar to the IFS (which also shows more drizzle bias)}
  \label{fig:quantiles}
\end{figure}

In Figure \ref{fig:quantiles}, we plot the fractional contribution \citep{Klingaman17, Leung22} of binned precipitation rates at 18hr lead time to the total precipitation in each gridbox over all initial conditions in 2018. This assesses how much of the total precipitation comes from light versus heavy precipitation, and we find that our GAN model achieves the best agreement with the ground truth in general. Notably, the IFS model exhibits more drizzle bias \citep{Chen21} for rates between 0.1-10 mm/6hrs, though we find that both IFS and our model overestimate the central mode of the fractional distribution at $\sim$5 mm/6hrs.

\begin{figure}[H]
  \centering
  \includegraphics[width=\linewidth]{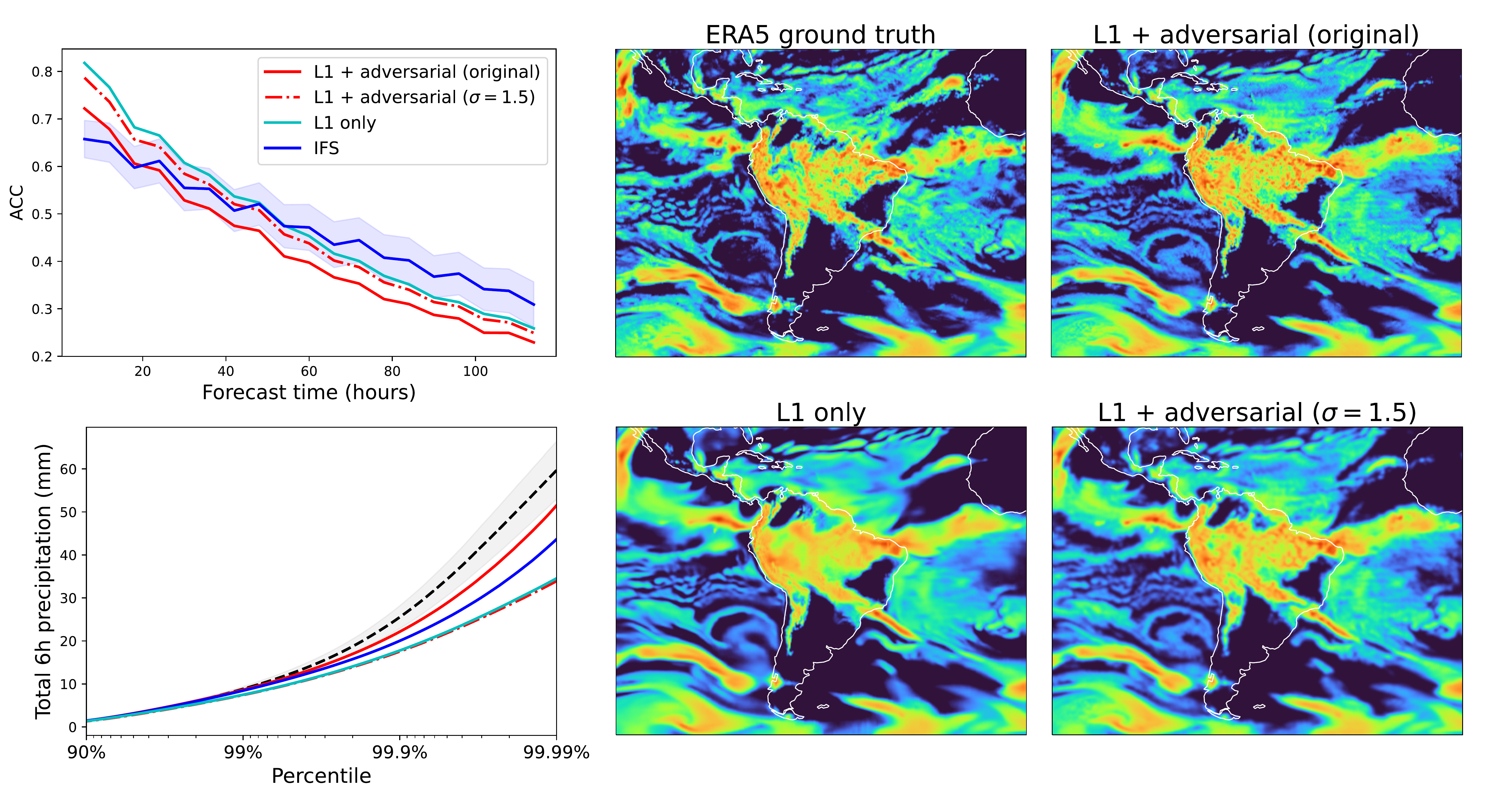}
  \caption{Applying a small amount of gaussian blurring ($\sigma=1.5$) to the outputs from the \emph{adversarial$+\mathcal{L}_1$} model leads to increased ACC over the entire forecast sequence as compared to the unaltered outputs (upper left panel). At the same time, the ability to capture extremes is degraded, as shown in the lower left panel. ACC and percentiles are averaged over 178 initial conditions, as in Figure \ref{fig:stats}, with variability over ACC additionally included for IFS.}
  \label{fig:blur}
\end{figure}

Finally, in Figure \ref{fig:blur}, we demonstrate tradeoffs between global metrics like ACC and quantifying extreme events. To do so, we conduct a simple experiment---we add gaussian blurring to the total precipitation outputs after inference and re-compute two metrics: ACC and extreme percentiles. While smoothing predictably leads to worse performance in capturing extremes, we observe that it does lead to higher ACC values. This could possibly be because ACC is a global metric and smoothing can help correct highly localized errors---this can inflate the quality of the forecast if we only focus on global behaviour. Hence, identifying the right metrics to characterize precipitations is very important. We leave evaluating our forecasts with a comprehensive set of diagnostic tools that focus on different aspects of modeling precipitation to future work.

\end{appendices}

\end{document}